\def\year{2021}\relax
\documentclass[letterpaper]{article} 
\usepackage{aaai21}  
\usepackage{times}  
\usepackage{helvet} 
\usepackage{courier}  
\usepackage[hyphens]{url}  
\usepackage{graphicx} 
\usepackage{graphicx}  
\usepackage{natbib}  
\usepackage{caption}
\usepackage{mathtools}
\usepackage{amsthm}
\usepackage{subcaption}
\usepackage{paralist}
\usepackage{bm}
\usepackage{amsfonts}
\usepackage{multirow}
\usepackage{cases}
\usepackage{booktabs}
\usepackage{threeparttable}
\usepackage{epstopdf}
\usepackage{upgreek}
\usepackage{endnotes}
\usepackage{etoolbox}
\usepackage{algpseudocode}
\usepackage{amsmath}
\usepackage{algorithm}
\usepackage{color}
\usepackage[switch]{lineno}

\title{Generalizable Representation Learning for Mixture Domain Face Anti-Spoofing}
\author{Zhihong Chen\textsuperscript{\rm 1,\rm 2}\footnote{Equal contribution.}, Taiping  Yao\textsuperscript{\rm 2}\footnotemark[\value{footnote}], Kekai Sheng\textsuperscript{\rm 2}, Shouhong Ding\textsuperscript{\rm 2}\thanks{Corresponding author.}, Ying Tai\textsuperscript{\rm 2}, \\ Jilin Li\textsuperscript{\rm 2}, Feiyue Huang\textsuperscript{\rm 2}, Xinyu Jin\textsuperscript{\rm 1}\\}
\affiliations{\textsuperscript{\rm 1}College of Information Science \& Electronic Engineering, Zhejiang University, \\ \textsuperscript{\rm 2} Youtu Lab, Tencent\\
\{zhihongchen, jinxy\}@zju.edu.cn, \{taipingyao, ericshding, yingtai, jerolinli, garyhuang\}@tencent.com, Shengkekai\_D@163.come
}

\begin{document}
\maketitle
\begin{abstract}
Face anti-spoofing approach based on domain generalization (DG) has drawn growing attention due to its robustness for unseen scenarios. Existing DG methods assume that the domain label is known. However, in real-world applications, the collected dataset always contains mixture domains, where the domain label is unknown. In this case, most of existing methods may not work. Further, even if we can obtain the domain label as existing methods, we think this is just a sub-optimal partition. To overcome the limitation, we propose domain dynamic adjustment meta-learning (D$^2$AM) without using domain labels, which iteratively divides mixture domains via discriminative domain representation and trains a generalizable face anti-spoofing with meta-learning. Specifically, we design a domain feature based on Instance Normalization (IN)  and propose a domain representation learning module (DRLM) to extract discriminative domain features for clustering. Moreover, to reduce the side effect of outliers on clustering performance, we additionally utilize maximum mean discrepancy (MMD) to align the distribution of  sample features to a prior distribution, which improves the reliability of clustering. Extensive experiments show that the proposed method outperforms conventional DG-based face anti-spoofing methods, including those utilizing domain labels. Furthermore, we enhance the interpretability through visualization.
\end{abstract}

\section{Introduction} \label{introduction}
Despite recent significant progress, the security of face recognition systems is still vulnerable to presentation attacks (PA), \textit{e.g.,} photo, video replay, or 3D facial mask. To cope with these presentation attacks,  face anti-spoofing \cite{tan2010face,liu2018learning} is deployed as a pre-step prior to face recognition. Various face anti-spoofing methods have been proposed,  which assume that there are inherent differences between live and spoof faces, such as color textures \cite{boulkenafet2016face}, image distortion cues \cite{wen2015face}, temporal variation \cite{shao2017deep}, or deep features \cite{yang2014learn,zhang2020face}. Although these methods achieve promising performance in intra-dataset experiments,  the performance dramatically degrades under a cross-domain dataset. To improve generalization ability under unseen situations, a variety of domain generalization (DG)-based methods \cite{wang2020cross,shao2020regularized,qin2019learning,jia2020single,saha2020domain} are proposed by leveraging domain labels from multiple source domains.
\begin{figure}[!t]
   \centering
     \includegraphics*[width=0.95\linewidth]{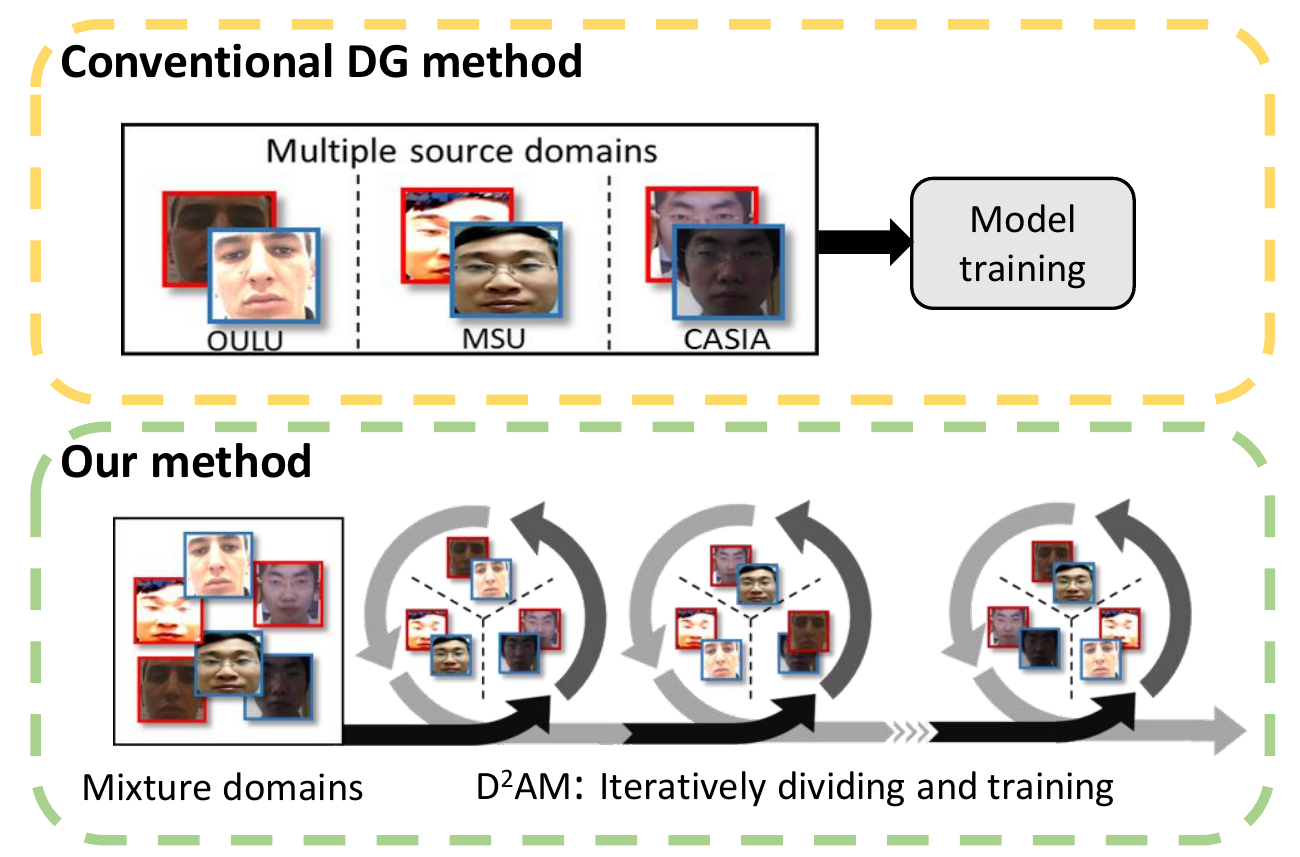}
   \caption{Unlike conventional DG method that requires domain labels, our method can iteratively assign pseudo domain labels and be trained using meta-learning by D$^2$AM.}
   \label{Fig1}
\end{figure}

These DG-based methods require domain labels that indicate where each sample in multiple source domains comes from.  However, in a more practical scenario, we may obtain a mixture domain dataset, in which the domain label of each sample is unknown as shown in Figure \ref{Fig1}. The existing DG methods may not work if the domain label is not available. There are mainly three challenges to relax the constraint: (1) While it could be solved by assigning the domain label manually, it can be very expensive and time-consuming; (2) Even worse, since the domain information of face acti-spoofing is composed of various factors such as illumination, background, camera type, \textit{etc.,} it is not clear how to define the domain and divide the mixture domains; (3) Further, even if we can obtain the domain label which defined as the dataset to which the sample belongs, as existing methods, we think this is just a suboptimal partition. The reason is that the samples among different datasets may partially overlap in distribution, especially when the model can extract a certain degree of domain invariant features in the middle stage of training, which may cause the model to be unable to focus on better-generalized learning directions due to the small distribution difference between domains.

To relax the constraint that DG-based methods need domain labels, and ensure more difficult and abundant domain differences, as shown in Figure \ref{Fig1}, we propose a generalizable face anti-spoofing method, named \textbf{d}omain \textbf{d}ynamic \textbf{a}djustment \textbf{m}eta-learning (D$^2$AM), which iteratively divides mixture domains via discriminative domain representation for meta-learning. Specifically, to define the domain information, considering that Instance Normalization (IN) in the networks can alleviate the domain discrepancy \cite{zhou2019omni},  we exploit a stack of convolutional feature statistics (\textit{i.e.,} mean and standard deviation) to get domain representation. To extract the domain-discriminative feature, we design a  domain representation learning module (DRLM) to extract discriminative domain features under the guidance of the channel attention mechanism. Further, a domain enhancement entropy loss is added to DRLM to enhance the confusion of task discrimination information in domain channels. Once discriminative domain representation is obtained, our method iteratively assigns pseudo domain labels by clustering, and trains a domain-invariant feature extractor by meta-learning. In addition, to prevent outliers from affecting the performance of clustering, we introduce an MMD-based regularization in the adaptation layer, \textit{i.e.,} the previous layer of the output layer, to reduce the distance between the sample feature distribution and the prior distribution. Furthermore, the embedding of MMD-based regularization can encourage the model to learn to correct the distribution of unseen samples through meta-learning.

The main contributions of this work are summarized as follows: (1) We propose a novel and realistic mixture domain face anti-spoofing scenario and design \textbf{d}omain \textbf{d}ynamic \textbf{a}djustment \textbf{m}eta-learning (D$^2$AM) to address this scenario. (2) Domain representation learning module (DRLM) and  MMD-based regularization are designed for better dynamic adjustment. (3) Extensive experiments and visualizations are presented, which demonstrates the effectiveness of D$^2$AM against the state-of-the-art competitors.
\section{Related Work}
\subsubsection{Face Anti-Spoofing} Recent face anti-spoofing approaches can be roughly classified into three categories: conventional approaches, deep learning approaches, and domain generalized approaches. Conventional approaches detect attacks by texture cues, which adopt hand-craft features to differentiate real/fake faces, such as LBP \cite{de2014face}, HOG \cite{gragnaniello2015investigation}, SURF \cite{boulkenafet2016face}, SIFT \cite{patel2016secure}, \textit{etc}.  With the recent success of deep learning in computer vision, various deep methods are employed. In \cite{yang2014learn},  discriminative deep features are extracted by CNN for real/fake faces classification.  Liu \textit{et al.} \cite{liu2018learning} propose a CNN-RNN architecture, which leverages face depth and rPPG signal estimation as auxiliary supervision to assist in attacks detect. And the work in \cite{jourabloo2018face} inversely decompose a spoof face into a live face and a noise of spoof for classification. Although these methods work well under intra-dataset scenarios, their performance becomes degraded in unseen scenarios. In light of this,  some domain generalized approaches are proposed. Shao \textit{et al.} \cite{shao2019multi} propose to learn domain-invariant representation by multi-adversarial deep domain generalization for face anti-spoofing, while  Jia \textit{et al.} \cite{jia2020single} design the single-side adversarial learning and the asymmetric triplet loss to further improve the performance. The most related work to ours is proposed in \cite{shao2020regularized}, where meta-learning is explored with domain knowledge for generalizable face anti-spoofing. However, this method requires domain labels, which are not satisfied in the novel scenario, \textit{i.e.,} mixture domain face anti-spoofing. 

\subsubsection{Deep Domain Generalization}
 Several deep DG methods have been proposed. Ghifary \textit{et al.} \cite{ghifary2015domain} match the feature distributions among multiple source domains by using an auto-encoder. The work in \cite{li2018domain} aligns multiple domains to a pre-defined distribution via adversarial learning. MLDG \cite{li2018learning} designs a model-agnostic meta-learning for DG.  Note that,  the work \cite{matsuura2020domain} has achieved DG by clustering a mixture of multiple latent domains. However, the domain features they extracted may contain task discriminative information because they did not consider distilling that information, which may hinder the domain clustering.  While our method designs a DRLM with style enhancement entropy loss, which encourages model to extract domain-discriminative features for better domain dividing.
\begin{figure*}[ht]
   \centering
     \includegraphics*[width=0.9\linewidth]{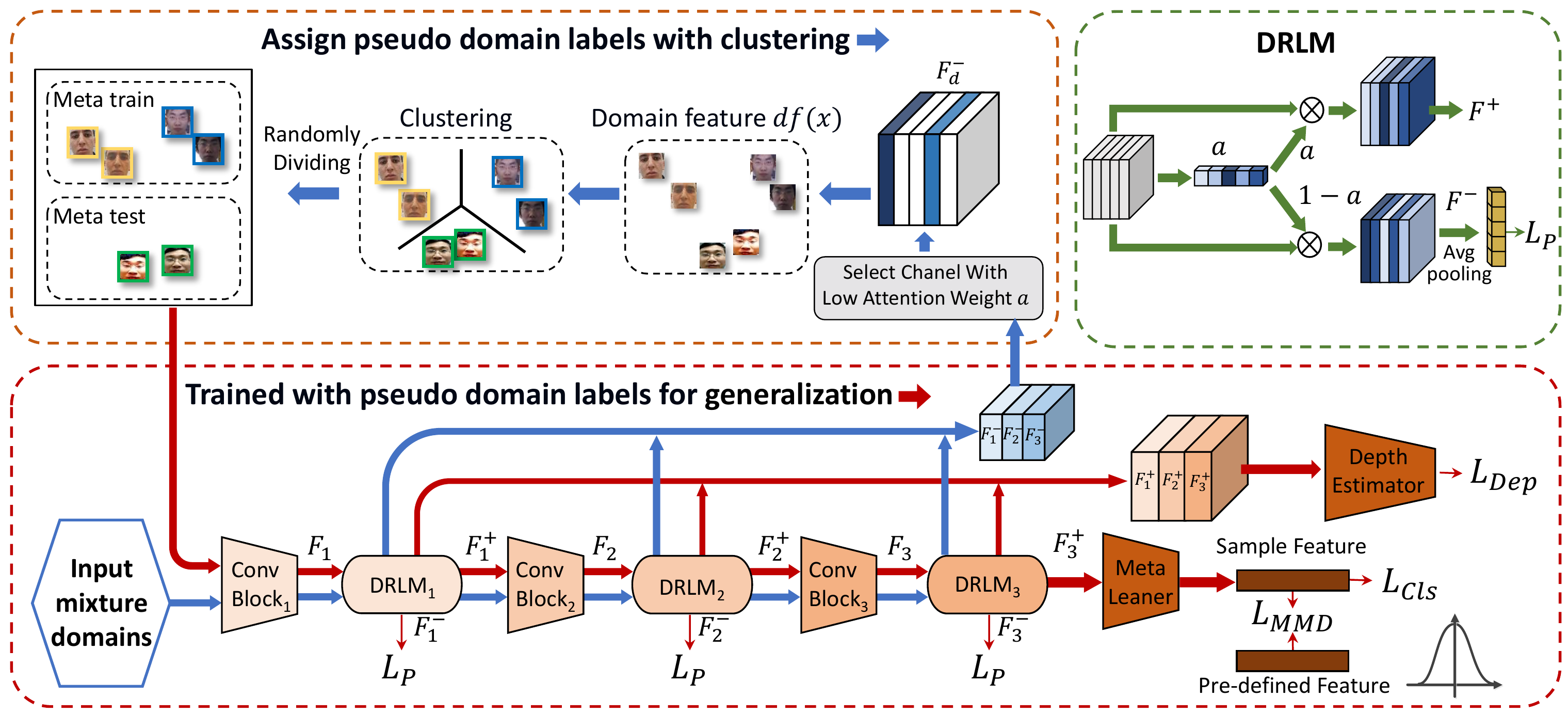}
     \caption{Overview of our method. Each epoch consists of two stages. In the first stage (blue flow), the pseudo domain label is assigned by clustering with discriminative domain representation. In the second stage (red flow), we train a face anti-spoofing model with meta-learning based on pseudo domain label. The green box is DRLM, which utilizes a Squeeze-and-Excitation (SE)-like framework to extract task-discriminative features $F^+$ and domain-discriminative features $F^-$.}
     \label{Fig2}
\end{figure*}
\section{Our Approach} \label{section3}
\subsubsection{Problem Definition and Notations}
Conventional  generalizable face anti-spoofing methods train the model that accurately works for the unseen domain $\mathcal{D}_t$  by using multiple source domains $\mathcal{D}_{ms}=\{(x_i^s,y_i^s,d_i^s)\}_{i=1}^{n_{s}}$,  where $x_i^s$ is the input image, $y_i^s$ is the task label and $d_i^s$ is the domain label. However, as mentioned above, the dataset may be a mixture of multiple latent domains, in which case it is difficult to get domain labels manually. Our goal is to improve generalization of the model through a mixture domain dataset $\mathcal{D}_{s}=\{(x_i^s,y_i^s)\}_{i=1}^{n_s}$ without the domain label $d_i^s$.

 \subsubsection{Overiew of D$^2$AM}
 Figure \ref{Fig2} shows the overall flowchart of our framework. Our method iteratively clusters domain-discriminative representation to reassign pseudo domain label to each sample, which can achieve simultaneously more abundant and more difficult domain shift scenarios for meta-learning.  In summary, in each epoch, our method consists of two stages. \textit{In the first stage} (blue flow), the pseudo domain label is assigned by clustering with discriminative domain representation. Specifically, due to the  clustering features extracted directly from the network without processing contain more task discrimination information, we process the features extracted by the model based on IN and convert them into domain features.  Moreover, we design the DRLM module with domain enhancement entropy loss to encourage the model to extract discriminative domain features that without task discrimination information, so as to avoid the model's ability to extract domain invariant features from hindering the extraction of domain-discriminative features.  \textit{In the second stage} (red flow), we train a face anti-spoofing model with meta-learning based on pseudo domain label. We also incorporate an MMD-based regularization into feature learning process,  which regularizes the feature space for better clustering and promotes the model to correct the distribution of unseen samples. The whole process is summarized in Algorithm \ref{alg::algorithm1}, and details are described as follows.
\subsection{Dynamically Assigning Pseudo Domain Labels}
\subsubsection{Domain Feature Define}
 We assume the domain information of an image can be represented by its style. IN performs a form of style normalization by normalizing feature statistics \cite{huang2017arbitrary}, which can be formed as:
\begin{equation}
\begin{split}
\label{e1}
IN(F)=\gamma(\frac{F-\mu(F)}{\sigma(F)})+\eta,
\end{split}
\end{equation}
where $F\in\mathbb{R}^{H\times W\times C}$ is the convolutional feature and  $H, W, C$ denote the height, width, and number of channels, respectively, $\gamma,\eta\in\mathbb{R}^C$ are learnable parameters, mean $\mu(\cdot)\in\mathbb{R}^C$ and standard deviation $\sigma(\cdot)\in\mathbb{R}^C$  are computed across spatial dimensions independently for each channel.
\begin{small}
\begin{gather}
\label{e2}
\mu_c(F)=\frac{1}{HW}\sum_{h=1}^{H}\sum_{w=1}^{W}(F_{hwc}), \\
\label{e3}
\sigma_c(F)=\sqrt{\frac{1}{HW}\sum_{h=1}^{H}\sum_{w=1}^{W}(F_{hwc}-\mu_c(F))^2+\epsilon}.
\end{gather}
\end{small}

Since the normalization by $\mu$ and $\sigma$ can alleviate domain discrepancy, we exploit a stack of convolutional feature statistics getting from multiple layers of the feature extractor to represent domain features. Hence, the domain feature can be defined as: $df(x)=\{\mu(F_1),\sigma(F_1),...,\mu(F_M),\sigma(F_M)\}$, where $M$ represents the $M$th convolutional layer.

\subsubsection{Domain Representation Learning Module} To obtain more discriminative domain features $df(x)$, we design the DRLM to extract domain-discriminative convolutional feature. Specifically, as shown Figure \ref{Fig2}, it is a SE-like framework, which is expected that channels with high attention contain more \textit{task information}, while channels with low attention contain more \textit{domain information}. Therefore, the module can be used to extract the task-discriminative feature $F^+$ and domain-discriminative feature $F^-$ by:
\begin{small}
\begin{gather}
\label{e4}
\bm{a}=Sigmoid(W_2ReLu(W_1pool(F))), \\
\label{e5}
F^+=\bm{a}\cdot F, \\
\label{e6}
F^-=(1-\bm{a})\cdot F,
\end{gather}
\end{small}
where $F^+ \in \mathbb{R}^{H\times W\times C}$ and $F^- \in \mathbb{R}^{H\times W\times C}$ are extracted through masking $F$ by a learned channel attention vector $\bm{a}=[a_1,a_2....,a_C] \in \mathbb{R}^C$. $pool$ is a global pooling layer, $W_1 \in \mathbb{R}^{C\times (C/\tau)}$ and $W_2 \in \mathbb{R}^{(C/\tau)\times C}$ are trainable weights, and $\tau$ is the dimension reduction ratio.

Minimizing the entropy regularization $-p \log p$ favors a low-density separation between classes and increases task discrimination \cite{grandvalet2005semi,chen2020selective}. Hence, we utilize reverse entropy loss, named domain enhancement entropy loss, to regularize $F^-$ to discard task discrimination information for better clustering. The loss can be formed as:
\begin{equation}
\begin{split}
\label{e7}
L_p&=P(F^-)\log P(F^-), \\
P(F^-)&=Sigmoid(W_ppool(F^-)),
\end{split}
\end{equation}
where $P(\cdot)$ represents the probability of whether it is a positive sample, and $W_p \in \mathbb{R}^{C \times 1}$ is the trainable weights. Note that whether it is a positive sample or a negative sample, $L_p$ can constrain $P(\cdot)$ to around 0.5, which means the task discrimination information of $F^-$ may be discarded, so as not to hinder domain clustering.
\subsubsection{Clustering with Discriminative Domain Representation} For obtaining pseudo domain labels, since channels with low attention weights contain more domain information, we sort the attention weights in ascending order and select the channel features of $F^-$ corresponding to the first $C/2$ values to form the discriminative domain convolutional feature $F^-_{d}$. To further remove task discrimination information, after obtaining domain features $\{df(x_i)\}_{i=1}^{n_s}$ through $F^-_{d}$ for all samples, we perform $K$-means clustering \cite{macqueen1967some} on the positive and negative samples respectively, assign a domain label to each sample, and finally, combine the positive and negative samples with the same domain label into one domain. The problem here is that clustering cannot properly decide which domain label should be assigned to each cluster,  which may lead to a mismatch between positive and negative domains and negatively affects the training.

To solve this problem, in the first epoch, we use the ResNet \cite{he2016deep} pre-trained on ImageNet to extract domain features to divide all samples into $K$ domains for correct positive and negative domain matching. After that, in each epoch, positive and negative samples are clustered using domain-discriminative features of our face anti-spoofing model and then combined. Based on \cite{matsuura2020domain}, we use Kuhn-Munkres algorithm \cite{munkres1957algorithms} to ensure the reassigned pseudo domain labels are not shifted largely with those from the previous epoch.

\subsection{MMD-Based Regularized Meta-Learning}
After obtaining pseudo domain labels, we randomly use $K-1$ domains as the meta-train $D^s_i (i=1,...,K-1)$ and the remaining one domain as the meta-test $D^t$ in each iteration.

\subsubsection{Meta-Train} We sample the batches $B_i (i=1,...,K-1)$ in each meta-train domain, and perform cross-entropy classification in each meta-train domain as follows:
\begin{footnotesize}
\begin{equation}
\begin{split}
\label{e8}
\mathop{L_{Cls(B_i)}}\limits_{\theta_E,\theta_M}=\sum_{(x,y)\in B_i}y\log M(E(x))+(1-y)\log(1- M(E(x)),
\end{split}
\end{equation}
\end{footnotesize}
where $\theta_E$ and $\theta_M$ represent the parameters of the feature extractor and the meta learner. To avoid the harmful influence of outliers on clustering, we propose the MMD-based regularization to constrain the feature space, which narrows the distance between outliers and sample dense areas, and encourages the meta-learner to correct unseen samples distribution. Specifically, this regularization reduces the distance between the sample feature distribution and the prior distribution on the adaptation layer, \textit{i.e.,} the previous layer of the output layer in the meta leaner, which can be formed as:
\begin{equation}
\begin{split}
\label{e9}
\mathop{L_{MMD(B_i)}}\limits_{\theta_E,\theta_M}=\|\frac{1}{b_i}\sum_{j=1}^{b_i}\phi(\bm{h}_{s_j})-\frac{1}{b_i}\sum_{j=1}^{b_i}\phi(\bm{h}_{t_j})\|_\mathcal{H}^2,
\end{split}
\end{equation}
where $b_i$ is the batch size, $\phi$ is the kernel function, $\mathcal{H}$ is the reproducing kernel Hilbert space (RKHS), $\bm{h}_{s_j}$ is the sample feature output by adaptation layer, and $\bm{h}_{t_j}$ is the feature of the same dimension as $\bm{h}_{s_j}$, which randomly generated from prior distribution. In each meta-train domain, the inner-update of meta leaner's parameters can be calculated as $\theta_{M_i'}=\theta_M-\alpha\nabla_{\theta_M}(L_{Cls(B_i)}(\theta_E,\theta_M)+\lambda_m L_{MMD(B_i)}(\theta_E,\theta_M))$, $\lambda_m$ is the hyper-parameter. Meanwhile, we incorporate face depth maps as auxiliary information to guide learning of extractor, which can be formed as:
\begin{footnotesize}
\begin{equation}
\begin{split}
\label{e10}
\mathop{L_{Dep(B_i)}}\limits_{\theta_E,\theta_D}=\sum_{(x,y)\in B_i}\|D(E(x))-I\|^2,
\end{split}
\end{equation}
\end{footnotesize}
where $\theta_D$ is the parameter of depth estimator and $I$ is the face depth map of face image, which estimated by PRNet \cite{feng2018joint} for real face and set zeros for fake face.

\subsubsection{Meta-Test} We sample batch $B_t$ in the one remaining meta-test domain $D^t$. We encourage our face anti-spoofing model trained in each meta-train domain can simultaneously perform well on the unseen cross-domain meta-test domain. Hence, we calculate $\Sigma_{i=1}^{K-1}(L_{Cls(B_t)}(\theta_E,\theta_{M_i'})+\lambda_m L_{MMD(B_t)}(\theta_E,\theta_{M_i'}))$ with inner-updated meta learners. Also, $L_{Dep(B_t)}(\theta_E,\theta_D)$ is incorporated like meta-train.
\begin{algorithm}[t]
\caption{The optimization strategy of our D$^2$AM}
\label{alg::algorithm1}
\begin{algorithmic}[1]
\State \textbf{Input:} Mixture domain dataset $\mathcal{D}_{s}=\{(x_i^s,y_i^s)\}_{i=1}^{n_s}$
\State Initialize model parameters $\theta_E$, $\theta_M$, $\theta_D$, and determine $K$ by pre-clustering with ResNet
\State\textbf{while} not end of epoch \textbf{do}
\State \quad \textbf{if} epoch==1
\State \qquad Calculate $\{df(x_i)\}_{i=1}^{n_s}$ using pre-trained ResNet
\State \quad \textbf{else}
\State \qquad Calculate $\{df(x_i)\}_{i=1}^{n_s}$ using feature extractor $E$
\State \quad Obtain $\{d_i\}_{i=1}^{n_s}$ by clustering  $\{df(x_i)\}_{i=1}^{n_s}$
\State \quad \textbf{while} not end of minibatch \textbf{do}
\State \qquad  Randomly use $K-1$ domains as the meta-train and the remaining one domain as the meta-test
\State \qquad  Meta-train: Sample the batches $B_i (i=1,...,K-1)$ in each meta-train domain
\State \qquad  \textbf{for} each batch $B_i$ \textbf{do}
\State \qquad \quad Calculate $L_{Cls(B_i)}(\theta_E,\theta_M)$, $L_{Dep(B_i)}(\theta_E,\theta_D)$,  $L_{MMD(B_i)}(\theta_E,\theta_M)$ and $L_{P(B_i)}(\theta_E)$ as Eq. \ref{e8}, \ref{e9}, \ref{e10}, \ref{e7}, respectively
\State \qquad \quad Inner update $\theta_{M_i'}$ with $L_{Cls(B_i)}(\theta_E,\theta_M)$, $L_{MMD(B_i)}(\theta_E,\theta_M)$
\State \qquad  Meta-test: Sample the meta-test batches $B_t$
\State \qquad  Use $\theta_{M_i', E, D}$ to calculate $L_{Dep(B_t)}(\theta_E,\theta_D)$, $\Sigma_{i=1}^{K-1}(L_{Cls(B_t)}(\theta_E,\theta_{M_i'})+\lambda_m L_{MMD(B_t)}(\theta_E,\theta_{M_i'}))$
\State \qquad  Meta-optimization with Eq. \ref{e11}, \ref{e12}, \ref{e13}
\State \textbf{Return:} Model parameters $\theta_E$, $\theta_M$, $\theta_D$
\end{algorithmic}
\end{algorithm}
\begin{figure*}[ht]
   \centering
     \includegraphics*[width=0.92\linewidth]{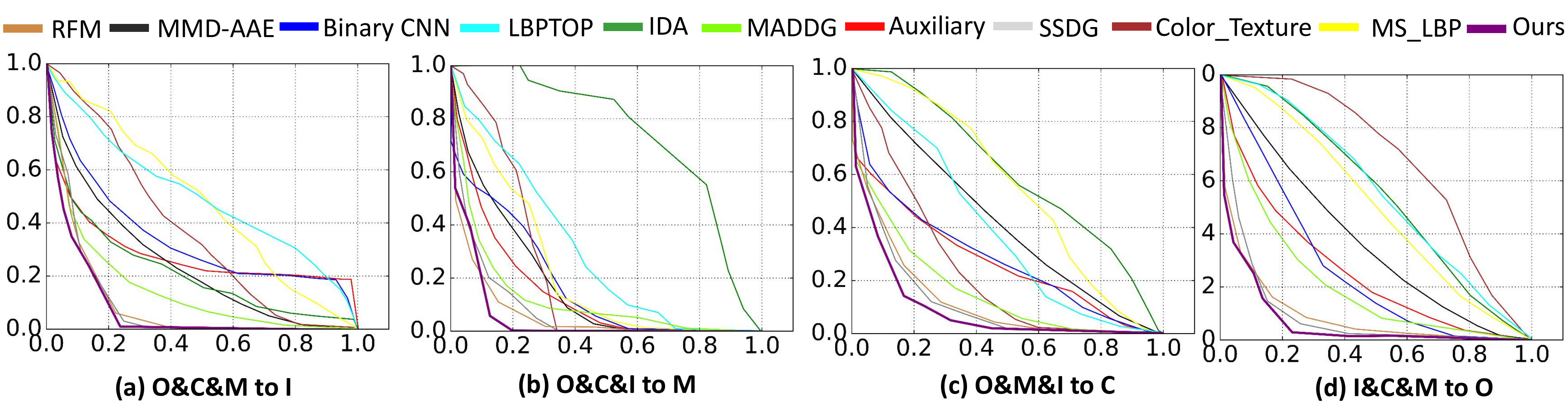}
     \caption{ROC curves of four testing sets for domain generalization on face anti-spoofing.}
\label{roc}
\end{figure*}
\begin{table*}[h]
  \centering
   \resizebox{\linewidth}{!}{
    \begin{tabular}{ccccccccc}
   \toprule
   \multirow{2}{*}{\textbf{Method}}&
    \multicolumn{2}{c}{\textbf{O\&C\&M to I}}&\multicolumn{2}{c}{\textbf{O\&C\&I to M}}&\multicolumn{2}{c}{\textbf{O\&M\&I to C}}&\multicolumn{2}{c}{\textbf{I\&C\&M to O}}\cr
    \cmidrule(lr){2-3} \cmidrule(lr){4-5} \cmidrule(lr){6-7} \cmidrule(lr){8-9}
    &HTER(\%)&AUC(\%)&HTER(\%)&AUC(\%)&HTER(\%)&AUC(\%)&HTER(\%)&AUC(\%)\cr
    \midrule
    MS\_LBP &50.30&51.64&29.76&78.50&54.28&44.98&50.29&49.31\cr
    Binary CNN   &34.47&65.88&29.25&82.87&34.88&71.94&29.61&77.54\cr
    IDA &28.35&78.25&66.67&27.86&55.17&39.05&54.20&44.59 \cr
    Color Texture  &40.40&62.78&28.09&78.47&30.58&76.89&63.59&32.71 \cr
    LBPTOP  &49.45&49.54&36.90&70.80&42.60&61.05&53.15&44.09 \cr
    Auxiliary (Depth)  &29.14&71.69&22.72&85.88&33.52&73.15&30.17&77.61 \cr
    Auxiliary (All)  &27.6&-&-&-&28.4&-&-&-\cr
    MMD-AAE  &31.58&75.18&27.08&83.19&44.59&58.29&40.98&63.08 \cr
    MADDG  &22.19&84.99&17.69&88.06&24.5&84.51&27.98&80.02 \cr
    SSDG-M  &18.21&90.61&16.67&90.47&23.11&85.45&25.17&81.83\cr
    RFM  &17.30&90.48&13.89&93.98&\textbf{20.27}&\textbf{88.16}&16.45&\textbf{91.16} \cr
    \midrule
    \textbf{D$^2$AM}&\textbf{15.43}&\textbf{91.22}&\textbf{12.70}&\textbf{95.66}&20.98&85.58&\textbf{15.27}&90.87\cr
    \bottomrule
    \end{tabular}}
     \caption{Comparison to face anti-spoofing methods on four testing sets for domain generalization on face anti-spoofing.}
    \label{tab:anti_1}
\end{table*}
\subsubsection{Meta-Optimization} We jointly train the three modules in our network in a meta-learning framework as follows:
\begin{footnotesize}
\begin{equation}
\begin{split}
\label{e11}
\theta_M&\leftarrow\theta_M-\beta\nabla_{\theta_M}(\sum_{i=1}^{K-1}(\mathop{L_{Cls(B_i)}}\limits_{\theta_E,\theta_M}+\lambda_m \mathop{L_{MMD(B_i)}}\limits_{\theta_E,\theta_M}\\
&+\mathop{L_{Cls(B_t)}}\limits_{\theta_E,\theta_{M_i'}}+\lambda_m \mathop{L_{MMD(B_t)}}\limits_{\theta_E,\theta_{M_i'}})),
\end{split}
\end{equation}
\end{footnotesize}
\begin{footnotesize}
\begin{equation}
\begin{split}
\label{e12}
\theta_E\leftarrow\theta_E-\beta\nabla_{\theta_E}(\mathop{L_{Dep(B_t)}}\limits_{\theta_E,\theta_D}+\sum_{i=1}^{K-1}(\mathop{L_{Cls(B_i)}}\limits_{\theta_E,\theta_M}+\lambda_m \mathop{L_{MMD(B_i)}}\limits_{\theta_E,\theta_M}\\
+\mathop{L_{Cls(B_t)}}\limits_{\theta_E,\theta_{M_i'}}+\lambda_m \mathop{L_{MMD(B_t)}}\limits_{\theta_E,\theta_{M_i'}}+\sum_{j=1}^{3}\lambda_p\mathop{L_{P(B_i)}^j}\limits_{\theta_E}+\mathop{L_{Dep(B_i)}}\limits_{\theta_E,\theta_D})),
\end{split}
\end{equation}
\end{footnotesize}
\begin{footnotesize}
\begin{equation}
\begin{split}
\label{e13}
\theta_D\leftarrow\theta_D-\beta\nabla_{\theta_D}(\mathop{L_{Dep(B_t)}}\limits_{\theta_E,\theta_D}+\sum_{i=1}^{K-1}(\mathop{L_{Dep(B_i)}}\limits_{\theta_E,\theta_D})),
\end{split}
\end{equation}
\end{footnotesize}
where $L_{P(B_i)}^j$ denotes the domain enhancement entropy loss for the $j$th DRLM in feature extractor $E$. Since we iteratively re-assign the domain label with the largest domain shift to the sample by clustering with discriminative domain representation in each epoch, this can simulate more abundant and difficult cross-domain scenarios and make the model focus on better-generalized learning directions. The detailed training process is shown in Algorithm \ref{alg::algorithm1}.
\section{Experiments}\label{section4}
\subsubsection{Datasets} Four public face anti-spoofing datasets are utilized to evaluate the effectiveness of our method: OULU-NPU \cite{boulkenafet2017oulu} (denoted as O), CASIA-FASD \cite{zhang2012face} (denoted as C), Idiap Replay-Attack \cite{chingovska2012effectiveness} (denoted as I), and MSU-MFSD \cite{wen2015face} (denoted as M). We randomly select three datasets as a mixture source domain for training,  and the remaining one is the unseen domain for testing. Unlike existing methods that assume each dataset represents a domain, the source domain we select is a mixture of multiple latent domains without domain labels.

\subsubsection{Implementation Details}
Our method is implemented via PyTorch and trained with Adam optimizer. We extract the RGB and HSV channels of each input image, thus, the input size is $256\times256\times6$. The learning rates $\alpha, \beta$ are set as 1e-3, 1e-4, respectively, and the prior distribution for MMD is defined as the standard normal distribution. For other hyper-parameters, we set $\lambda_p$ as 0.1 and $\lambda_m$ as 0.05. In our method, $K$ determines the number of subdomains that the model needs to be divided. We found that converting the convolutional features extracted by pre-trained ResNet into domain features for clustering can clearly divide the sample into several clusters, so we can determine the value of $K$ as 3. We use the Half Total Error Rate (HTER) and the Area Under Curve (AUC) as the evaluation metrics.
\begin{table}[h]
  \centering
   \resizebox{\linewidth}{!}{
    \begin{tabular}{ccccc}
   \toprule
   \multirow{2}{*}{\textbf{Method}}&
    \multicolumn{2}{c}{\textbf{M\&I to C}}&\multicolumn{2}{c}{\textbf{M\&I to O}}\cr
    \cmidrule(lr){2-3} \cmidrule(lr){4-5}
    &HTER(\%)&AUC(\%)&HTER(\%)&AUC(\%)\cr
    \midrule
    MS\_LBP &51.16&52.09&43.63&58.07\cr
    IDA&45.16&58.80&54.52&42.17\cr
    CT&55.17&46.89&53.31&45.16 \cr
    LBPTOP&45.27&54.88&47.26&50.21 \cr
    MADDG&41.02&64.33&39.35&65.10 \cr
    SSDG-M&\textbf{31.89}&71.29&36.01&66.88 \cr
    RFM&36.34&67.52&29.12&72.61 \cr
    \midrule
    \textbf{D$^2$AM}&32.65&\textbf{72.04}&\textbf{27.70}&\textbf{75.36} \cr
    \bottomrule
    \end{tabular}}
    \caption{Comparison to methods with limited domains.}
     \label{tab:anti_3}
\end{table}
\begin{figure*}[ht]
    \centering
  \begin{subfigure}[b]{0.24\textwidth}
    \includegraphics[width=\textwidth]{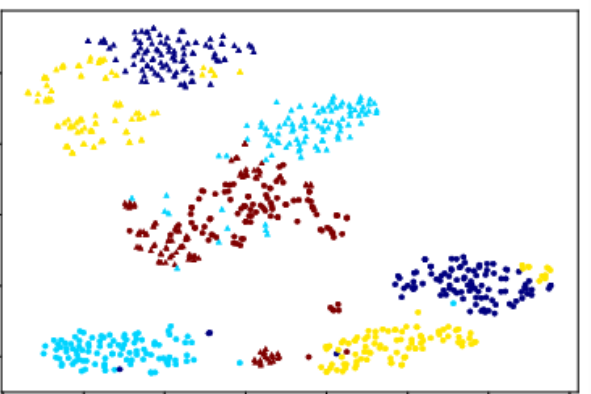}
    \caption{RFM}
    \label{2D1}
  \end{subfigure}
   \begin{subfigure}[b]{0.24\textwidth}
    \includegraphics[width=\textwidth]{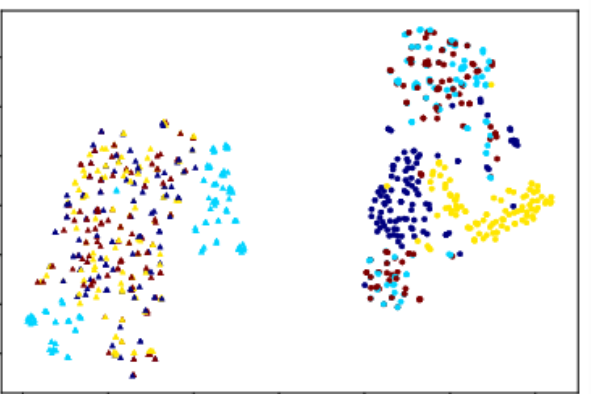}
    \caption{D$^2$AM}
    \label{2D2}
  \end{subfigure}
    \begin{subfigure}[b]{0.24\textwidth}
    \includegraphics[width=\textwidth]{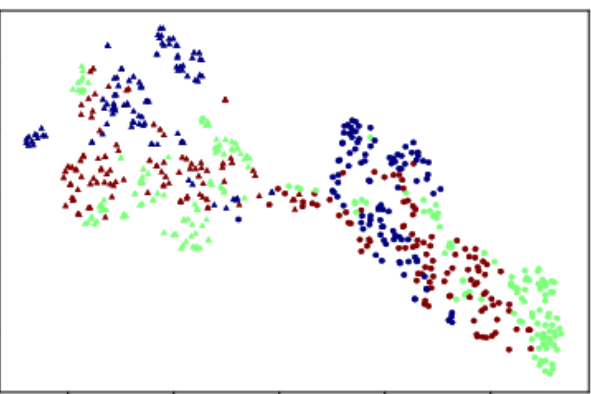}
    \caption{$F^+$}
    \label{2D3}
  \end{subfigure}
    \begin{subfigure}[b]{0.24\textwidth}
    \includegraphics[width=\textwidth]{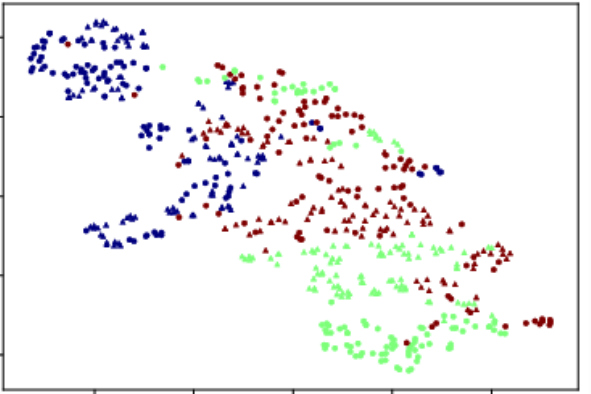}
    \caption{$F^-$}
    \label{2D4}
  \end{subfigure}
\caption{ The t-SNE visualization of the O\&C\&M to I task. Each color represents a domain (blue points in (a), (b) represent the unseen target domain),  square points and triangle points represent live faces and fake faces, respectively. Note that,  for better comparison, the domain label in the visualization is consistent with RFM, not the pseudo domain label assigned by D$^2$AM.}
\label{tsne}
\end{figure*}
\subsection{Result and Discussion}
As shown in Figure \ref{roc}, Tables \ref{tab:anti_1} and \ref{tab:anti_3},  our method outperforms all state-of-the-art methods under most of tasks.  We make the following observations from the results. (1) DG-based face anti-spoofing methods perform better than conventional methods. This proves that the distribution of target domain is different from source domain, while conventional methods only focus on the differentiation cues that only fit source domain. (2) The proposed method outperforms other DG-based methods, including RFM utilizing domain labels. We believe the reason is that our method can iteratively cluster mixture domains to find subdomains with the largest distribution difference, which allows the model to find better optimization directions with more abundant and difficult domain shift scenarios. Besides, we jump out of the box and evaluate it on single domain (protocol 3 of Oulu-NPU). The average results are: D$^2$AM (HTER=0.023$\pm$0.014, AUC=0.976$\pm$0.018) and RFM (HTER=0.031$\pm$0.016, AUC=0.947$\pm$0.025), which indicate that D$^2$AM has the strong ability to capture domain information.
\begin{table}[h]
  \centering
    \begin{tabular}{ccc}
   \toprule
   \textbf{Method}&HTER(\%)&AUC(\%)\cr
    \midrule
    D$^2$AM w/ d &18.24&89.28\cr
    D$^2$AM w/o d&27.05&73.57\cr
    D$^2$AM w/o select&16.57&89.98 \cr
    D$^2$AM w/o $L_p$&15.89&90.81\cr
    D$^2$AM w/o $L_{MMD}$&16.11&90.24 \cr
    \midrule   
    D$^2$AM($K$=2)&17.16&90.03 \cr
    D$^2$AM($K$=3)&\textbf{15.43}&\textbf{91.22} \cr
    D$^2$AM($K$=4)&16.82&90.57 \cr
    \bottomrule
    \end{tabular}
     \caption{Evaluations of different components of the proposed method on O\&C\&M to I task.}
  \label{tab:anti_2}
\end{table}
\subsubsection{Ablation Study}\label{Ablation Study}
We perform ablation study to verify the efficacy of each component. Several observations can be made from Table \ref{tab:anti_2}. (1) D$^2$AM w/o d means that D$^2$AM randomly selects meta-train and meta-test from mixture domains,  and its performance is worse than D$^2$AM w/ d, which utilizes domain label as existing methods. This is because there is no domain shift between the randomly selected domains. Hence, we think it is important to simulate difficult and abundant domain shift scenarios for meta-learning. (2) D$^2$AM w/o select means that features of all channels are used instead of only the channel features with low attention weight for clustering, which performs worse than D$^2$AM. These results indicate that features with low attention weights contain more domain information to achieve better clustering. (3) We conduct a sensitivity analysis for $K$, and the results show that determining $K$ based on the number of clusters divided by pre-trained ResNet can get the best results. (4) D$^2$AM yields the best performance, confirming that each component contributes to the final results.

\subsection{Visualization and Analysis}
\subsubsection{Adaptation Feature Visualization}
We visualize the features of adaptation layer using t-SNE \cite{donahue2014decaf}.  As shown in Figures \ref{2D1} and \ref{2D2}, several observations can be made. (1) From the perspective of domain information, comparing Figure \ref{2D1} and \ref{2D2}, we can find that D$^2$AM is more powerful to extract domain invariant features, which proves that our method is more robust to unseen samples. (2) From the perspective of task discriminative information, D$^2$AM makes features more dispersed in the feature space compared to RFM. Therefore, a better class boundary can be achieved by D$^2$AM. (3) There are no outliers in the feature space of D$^2$AM, which shows that MMD-based regularization can effectively constrain outliers to dense sample areas.
\subsubsection{Convolutional Feature Visualization}
In Figures \ref{2D3} and \ref{2D4}, we visualize the distribution of the features from the 3rd DRLM via t-SNE. It can be seen that $df(x)$ with $F^+$ contains more task discriminative information, while $df(x)$ with $F^-$ contains more domain information. This result validates that the domain enhancement entropy loss $L_p$ can encourage model to extract domain-discriminative features.
\begin{figure}[!t]
   \centering
     \includegraphics*[width=0.8\linewidth]{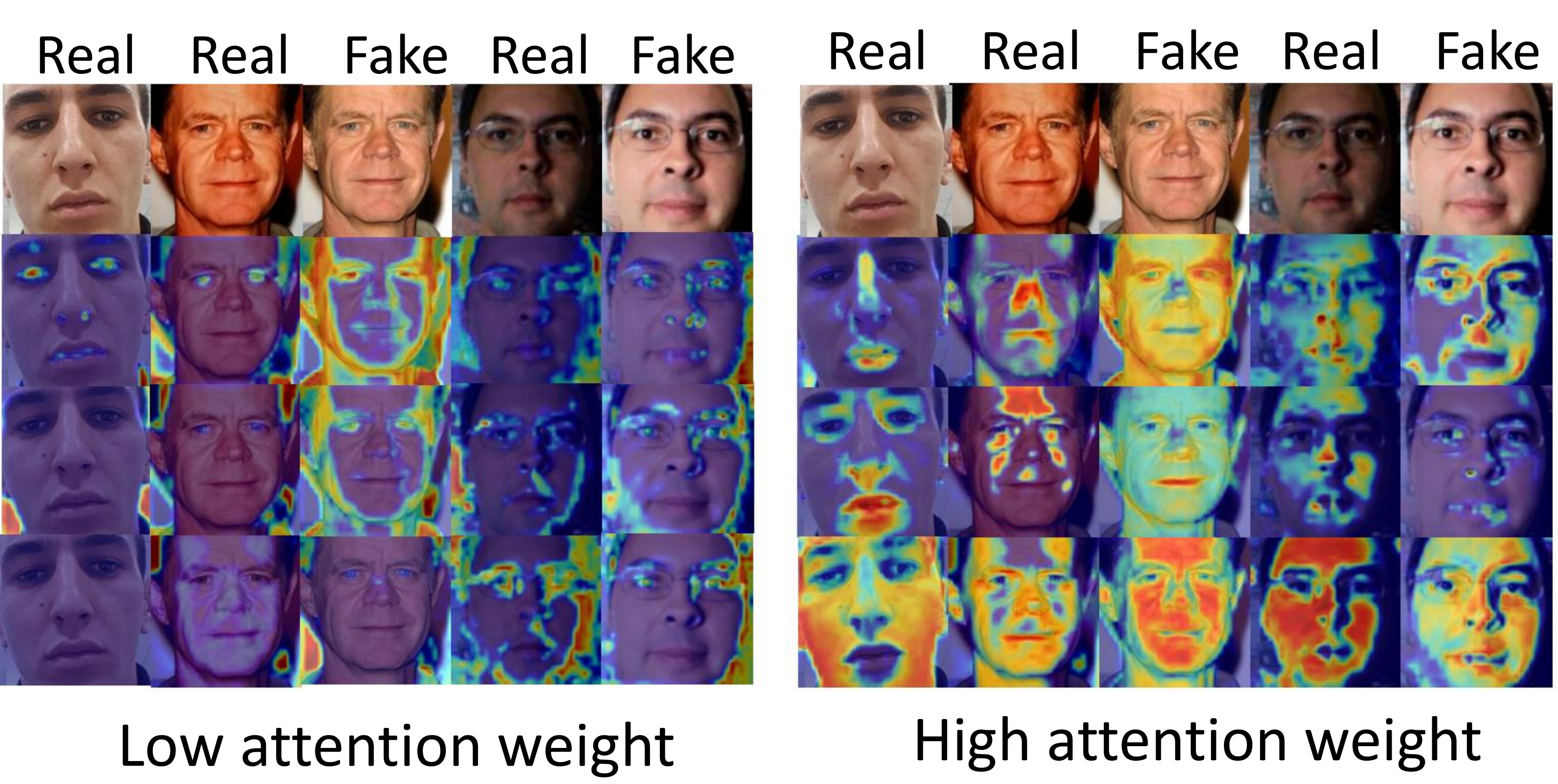}
   \caption{Attention map visualization for $F$.}
   \label{atten}
\end{figure}
\begin{figure}[!t]
   \centering
     \includegraphics*[width=0.75\linewidth]{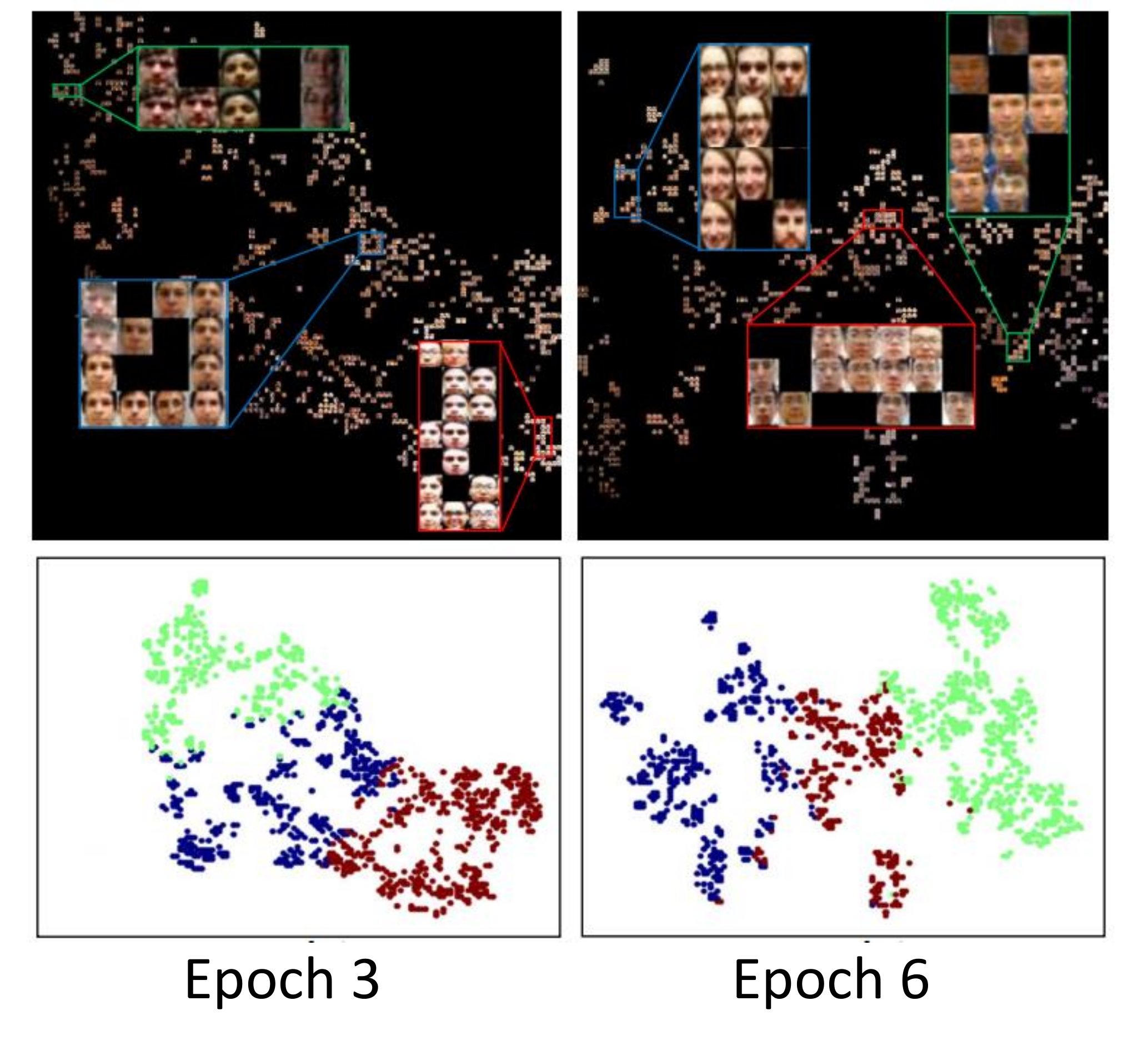}
   \caption{The t-SNE cluster visualization on O\&C\&M to I. Each color represents a domain divided by our method.}
   \label{face}
\end{figure}
\subsubsection{Attention Map Visualization}
To verify that channels with low attention weight in $F$ contain more domain information, we visualize attention maps of 3rd DRLM by the Global Average Pooling (GAP)  method \cite{zhou2016learning}.  As shown in Figure \ref{atten}, we find that attention maps with low attention weight focus on specific attack differentiation cue, backgrounds, overall styles, \textit{etc},  which are not generalized because they will be changed if data comes from a new scene. For example, in the second row of low attention weight, there is a type of attack that mimics eye blinking through two pieces of paper, so the type can be judged by locating a specific difference in the eye region, but other attacks do not have this difference, and in the third row of low attention weight, the background is focused.  While attention map with high attention weight always focuses on the region of the internal face, which are more likely to be intrinsic and generalized.  Therefore,  better domain dividing can be achieved by selecting features with low attention weight.
\subsubsection{Cluster Visualization}
To provide more insights on why our iterative clustering can perform better than other with known domain labels, we randomly sample 10,000 samples and cluster them according to $df(x)$ with $F_{d}^-$. The results are shown in Figure \ref{face}. We can find that D$^2$AM can dynamically adjust the basis of division, so as to simulate richer domain shifts for better meta-learning. For example, Epoch 3 is mainly clustered by illumination, while the blue points in the Epoch 6 contain samples of different illumination, so Epoch 6 is mainly clustered by background. The reason why our method can dynamically adjust is that meta-learning learns from the scene of current epoch to improve the robustness of this kind of shift, so in the next epoch, the model will extract domain invariant features for this scene, so as to guide clustering to focus on other domain information that is not yet robust and construct new scenarios.
\begin{figure}[!t]
   \centering
     \includegraphics*[width=0.85\linewidth]{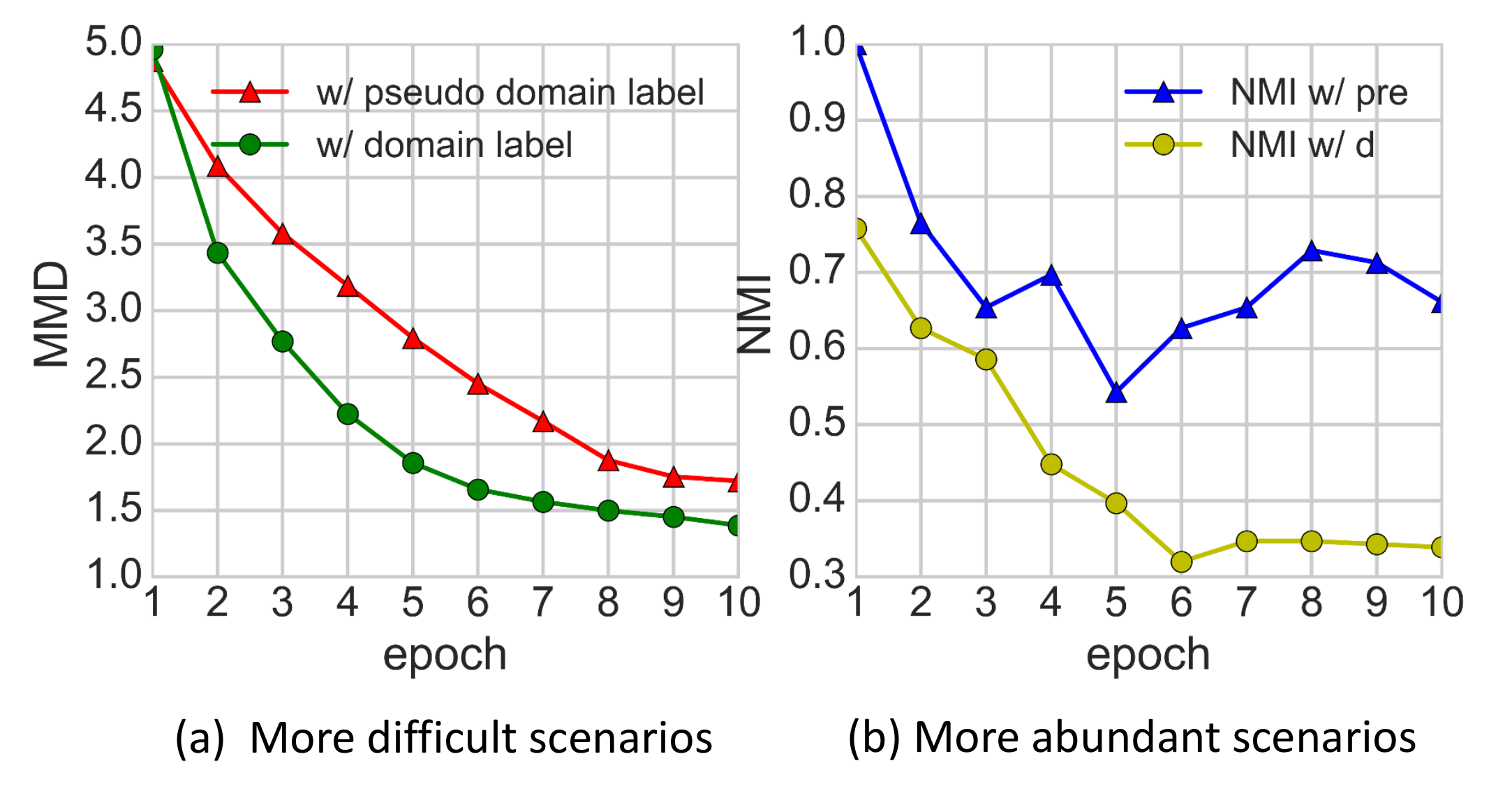}
   \caption{On O\&C\&M to I, (a) The MMD calculated with actual domains or pseudo domains. (b) NMI between pseudo domain and previous assignments or actual domain.}
   \label{MMD}
\end{figure}
\subsubsection{Cluster Analysis} To verify the effectiveness of our iterative clustering, we adopt MMD to calculate the difference between domains, and Normalized Mutual Information (NMI) \footnote{ The smaller the NMI, the greater the difference betwee them.} to measure the changes of pseudo-domain labels to evaluate the difficulty and diversity of domain shift scenarios, respectively. As shown in Figure \ref{MMD}a, compared with domain label which indicates the dataset each sample comes from, the subdomain obtained by pseudo domain label given by D$^2$AM has a larger domain difference, which shows that our method can simulate more difficult scenarios for better optimization direction. As shown in Figure \ref{MMD}b, we found that NMI between pseudo domain labels and previous assignments is almost between 0.6 and 0.8, indicating that the pseudo-domain label of the sample is constantly changing, especially in the middle stage of training, which indicates that D$^2$AM can generate richer domain shift scenarios to further improve the generalization. To measure the difference between the pseudo-domain and actual domain, we calculated the NMI between them and found that the pseudo-domain label and domain label only overlap about 64.9\%$\pm$5.7\%, which indicates that the pseudo-domain label is different from the actual domain label.
\section{Conclusion}
In this paper, to the best of our knowledge, this is the first wok to address mixture domain face anti-spoofing, where the domain label is unknown. Specifically, we design the D$^2$AM, which iteratively clusters mixture domains via discriminative domain representation and trains a generalizable feature extractor by meta-learning. A DRLM and  MMD-based regularization are designed for better dynamic adjustment to simulate more difficult and abundant domain shift scenes. Comprehensive experiments show that D$^2$AM outperforms conventional DG-based face anti-spoofing methods, including those utilizing domain labels. Furthermore, we enhance the interpretability through visualization.
\bibliography{anti_2021}
\end{document}